\documentclass{article}
\pdfoutput=1

\usepackage{arxiv}

\usepackage{microtype}
\usepackage{graphicx}
\usepackage{subfigure}
\usepackage{booktabs} 

\usepackage[utf8]{inputenc} 
\usepackage[T1]{fontenc}    
\usepackage{hyperref}       
\usepackage{url}            
\usepackage{booktabs}       
\usepackage{amsfonts}       
\usepackage{nicefrac}       
\usepackage{microtype}      
\usepackage{lipsum}

\usepackage{algpseudocode}
\usepackage{algorithm}

\usepackage{authblk}

\title{Neuromodulated Goal-Driven Perception in Uncertain Domains}

\author[1,*]{\textbf{Xinyun Zou}}
\author[2,$\dagger$]{\textbf{Soheil Kolouri}}
\author[2,$\ddagger$]{\textbf{Praveen K.~Pilly}}
\author[3,1,+]{\textbf{Jeffrey L.~Krichmar}}

\affil[1]{Department of Computer Science, University of California, Irvine, Irvine, CA, 92697, USA}
\affil[2]{Information and Systems Sciences Laboratory, HRL Laboratories, Malibu, CA, 90265, USA}
\affil[3]{Department of Cognitive Sciences, University of California, Irvine, Irvine, CA, 92697, USA}
\affil[*]{\texttt{xinyunz5@uci.edu}}
\affil[$\dagger$]{\texttt{skolouri@hrl.com}}
\affil[$\ddagger$]{\texttt{pkpilly@hrl.com}}
\affil[+]{\texttt{jkrichma@uci.edu}}

\begin{document}
\maketitle

\begin{abstract}
In uncertain domains, the goals are often unknown and need to be predicted by the organism or system. In this paper, contrastive excitation backprop (c-EB) was used in a goal-driven perception task with pairs of noisy MNIST digits, where the system had to increase attention to one of the two digits corresponding to a goal (i.e., even, odd, low value, or high value) and decrease attention to the distractor digit or noisy background pixels. Because the valid goal was unknown, an online learning model based on the cholinergic and noradrenergic neuromodulatory systems was used to predict a noisy goal (expected uncertainty) and re-adapt when the goal changed (unexpected uncertainty). This neurobiologically plausible model demonstrates how neuromodulatory systems can predict goals in uncertain domains and how attentional mechanisms can enhance the perception of that goal. 
\end{abstract}


\section{Introduction}
\label{intro}
Artificial top-down attentional systems tend to respond to sensory inputs similarly regardless of context and goals. However, biological systems select relevant information to guide behavior in the face of noisy and unreliable signals, as well as rapidly adapt to unforeseen situations. Goal-driven perception treats the same situation differently based on context and causes attention to be directed to goal-relevant inputs. Oftentimes, these goals are unknown and must be learned through experience. Moreover, these goals or contexts can shift without warning. Goal-driven perception helps prevent over-emphasizing less relevant stimuli and focus instead on critical stimuli that require an immediate response. 

Neuromodulators are important contributors for attention and goal-driven perception. In particular, the cholinergic system drives bottom-up, stimulus-driven attention, as well as top-down, goal-directed attention \cite{avery2014}. Furthermore, the cholinergic system increases attention to task-relevant stimuli, while decreasing attention to the distractions \cite{baxter1999cognitive,oros2014learning}. This is a similar idea to contrastive Excitatory Backpropagation (c-EB) where a top-down excitation mask increments attention to the target features and an inhibitory mask decrements attention to distractors \cite{zhang2018top}. The noradrenergic system responds to surprises or large violations of priors. When the noradrenergic system responds phasically, where the neural activity rapidly and transiently increases, it causes a network reset (e.g., re-initializing weights or connections) that allows rapid adaptation and relearning under novel conditions \cite{bouret2005network,Grella445}.

We modified a c-EB network for use in a goal-driven perception task, where the system had to increase attention to the intended goal object and decrease attention to the distractor. Specifically, we presented to the network pairs of noisy MNIST digits. One goal class was to attend to the digit based on its parity (i.e., even or odd goal), and the other goal class was to attend based on the magnitude of the digit (i.e., low-value or high-value goal). In addition, we added a neuromodulation model to the head of the network architecture that regulated goal selection. Similar to Yu and Dayan's model \cite{angela2005uncertainty} of the cholinergic and noradrenergic neuromodulatory systems, we framed the task as an attentional task where the goal (even, odd, low or high) had to be learned from experience (\textit{goal identity}) and the goal might be noisy and rewarded with some probability (\textit{goal validity}).

\section{Methods}
\label{method}

\subsection{Modification of c-EB}

Excitation Backpropagation (EB) was developed as a goal-driven attentional framework for of a CNN classifier based on a probabilistic Winner-Take-All (WTA) process \cite{zhang2018top}. It can visualize the features at each layer in the hierarchy that are relevant to a given output neuron. An important extension of EB was to have contrastive excitation backpropagation (c-EB), which discriminated the goal pixels from distractors by cancelling out common winner neurons for different goals and amplifying discriminative neurons for the target goal \cite{zhang2018top}. 

We extended the PyTorch \cite{paszke2017automatic} implementation of c-EB \cite{greydanus2018}, whereas the original code for c-EB \cite{zhang2018top} was written in Caffe \cite{jia2014caffe}. A pair of contrastive activation signals from the neuron of interest was backpropagated to the layer immediately below it. Then we added these two signals together before performing EB over the remaining layers, which generated the probability of each given feature to excite the goal neuron without being cancelled out by the inhibitory mask. 

In our goal-driven perception task, we constructed an architecture that had pairs of noisy MNIST digits \cite{lecun1998gradient} as the input in the forward pass (see Figure \ref{fig:architecture}). The system increased attention to the digit corresponding to the intended goal and decreased attention to the distractor digit. One goal was to attend to the digit based on its parity (i.e., either odd or even), the other  goal was to attend to the digit based on its magnitude (i.e., low values between 0 and 4 inclusively or high values between 5 and 9 inclusively). This resulted in two subgoals within a goal. After supervised training on noisy pairs generated from the MNIST training dataset, c-EB was applied to the top-down attentional process on the test pairs and driven by one of the four subgoals to excite only the pixels relevant to the goal digit.

\subsection{ACh and NE Neuromodulation}

\begin{algorithm}[ht]
\caption{ACh and NE Neuromodulation Process}
\label{alg:neuromod}
\begin{algorithmic}
\State {\bfseries Input:} $beta$, $ne\textrm{\_}reset$, $maxACh$, $ach\textrm{\_}correct$, 
\State $ach\textrm{\_}incorrect$, $ne\textrm{\_}correct$, $ne\textrm{\_}incorrect$, 
\State $all\textrm{\_}pairs$, $validity\textrm{\_}options$, $num\textrm{\_}switches$, 
\State $trial\textrm{\_}interval$, $trial\textrm{\_}range$
\State Initialize $ACh = \textrm{ones}(4)$
\State Initialize $NE = ne\textrm{\_}reset$
\For{$i=1$ {\bfseries to} $num\textrm{\_}switches$}
\State Randomly choose $majorGoal$.
\State Pick $minorGoal$ in the same goal class.
\State Randomly pick $validity$.
\State $trialLength = trial\textrm{\_}interval \pm trial\textrm{\_}range$
\For{$t=1$ {\bfseries to} $trialLength$}
\State Pick a new $test\textrm{\_}pair$ from $all\textrm{\_}pairs$.
\State Randomly pick $r$ between [0,1).
\If{$r<validity$}
\State $trueGoal = majorGoal$
\Else
\State $trueGoal = minorGoal$
\EndIf
\State Select $guessGoal$ from $\textrm{Softmax}(ACh, beta)$.
\State Get $goalDigit$ from $test\textrm{\_}pair$ with $trueGoal$.
\State Get $predDigit$ with c-EB directed by $guessGoal$.
\If{$predDigit$ and $guessGoal$ are correct}
\State Multiply $ACh[guessGoal]$ with $ach\textrm{\_}correct$.
\State $NE=NE\times ne\textrm{\_}correct$
\State Assure $ACh[guessGoal]$ below $maxACh$.
\State Assure $NE$ above $ne\textrm{\_}reset$.
\Else
\State Multiply $ACh[guessGoal]$ with $ach\textrm{\_}incorrect$.
\State $NE=NE\times ne\textrm{\_}incorrect$
\EndIf
\State $AChLevel = \textrm{mean}(ACh)$
\If{$NE > AChLevel / (0.5+AChLevel)$}
\State $ACh = \textrm{ones}(4)$
\State $NE = ne\textrm{\_}reset$
\EndIf
\EndFor
\EndFor
\end{algorithmic}
\end{algorithm}

For goal-driven perception, the network must predict a goal when goals are not known \textit{a priori}. Similar to Yu and Dayan's model \cite{angela2005uncertainty} of the cholinergic (ACh) and noradrenergic (NE) neuromodulatory systems, the goal target (odd, even, high-value or low-value) was rewarded with a probability (goal validity), but that goal would change periodically (goal identity). Cholinergic neurons tracked \textit{expected uncertainties} of the potential goals. Noradrenergic neurons track \textit{unexpected uncertainties}, and respond phasicially when a goal identity change is detected. When the noradrenergic system responded phasically, it caused a network reset by re-initializing the cholinergic neural activities, which allowed rapid adaptation and relearning under novel conditions.

Algorithm \ref{alg:neuromod} shows the logic of our ACh and NE neuromodulatory model. There were four ACh neurons, each neuron corresponding to a goal (i.e., even, odd, low-value, or high-value), and one NE neuron. One of the four attentional goal tasks was selected as the major goal. The true goal identity was set to either the major goal or the minor goal according to the goal validity (see Section \ref{sec:automatic} for details). The true goal digit was obtained from the labels of the test pair by using the true goal identity. The ACh neuromodulator went through a softmax function to select an action. This guessed goal activated two neurons related to the goal in the top layer of our network architecture (Figures \ref{fig:architecture}, \ref{fig:procedure}), which directed c-EB in the backward pass to activate the goal-relevant pixels in the test pair and then predicted the digit in the forward pass. If the prediction was correct (which means that the guessed goal identity matched the true goal identity and the predicted digit matched the true goal digit), the ACh level increased and the NE level decreased; otherwise, the ACh level decreased and the NE level increased. If the NE level was above a threshold (see \cite{angela2005uncertainty} and Algorithm \ref{alg:neuromod}), ACh and NE were reset to baseline levels (i.e., 1.0 for ACh and 0.25 for NE). We set the values of $ach\textrm{\_}correct$, $ach\textrm{\_}incorrect$, $ne\textrm{\_}correct$, and $ne\textrm{\_}incorrect$ to be 1.04, 0.99, 0.97, and 1.02 respectively. However, there was a wide range of parameter values that could be used to produce stable results.

\subsection{Experimental Setup}

\subsubsection{Network Architecture}
\label{sec:architecture}

\begin{figure}[ht]
\vskip 0.2in
\begin{center}
\centerline{\includegraphics[width=0.8\columnwidth]{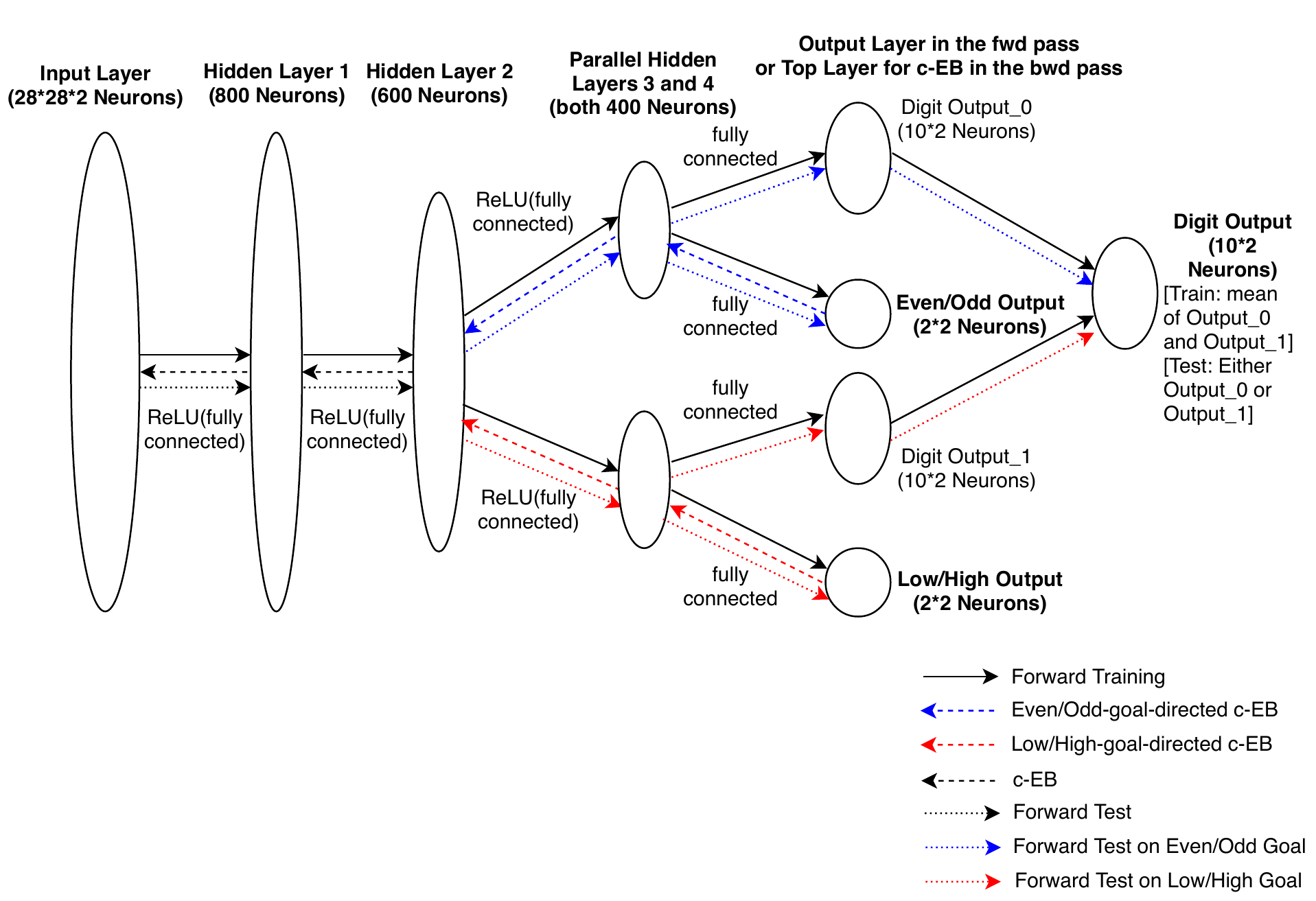}}
\caption{Network setup for our bottom-up classification process and our top-down attentional search process, with a pair of noisy MNIST digits as the input data in the forward pass. }
\label{fig:architecture}
\end{center}
\vskip -0.2in
\end{figure}

Figure \ref{fig:architecture} shows our bottom-up classification process and our top-down attentional search process. In the forward pass, the input layer received a pair of 28$\times$28 pixel MNIST digits \cite{lecun1998gradient} with extra noise and thus had $28\times 28\times 2=1568$ neurons. The noise, added to normalized pixel values (between 0 and 1) of the original MNIST digits, was randomly set between 0 and 0.7. There were two sequential fully-connected hidden layers, the first with 800 neurons and the second with 600 neurons. Next, there were two parallel fully-connected hidden layers, each with 400 neurons. One of them led to $2\times 2=4$ parity (even/odd) output neurons and $2\times 10=20$ digit output neurons. The other led to $2\times 2=4$ high-low (0-4 or 5-9) output neurons and 2x10=20 digit output neurons. During training, the final digit output took the average of the digit output generated from the two parallel hidden layers. All neurons in these layers implemented a ReLU activation function \cite{nair2010rectified}. 

The network was trained with pairs of noisy MNIST digits to learn the digits and their parity (even/odd) and magnitude (low/high-value) goal classes. Then during testing, one of the even, odd, high-value and low-value goal neurons was picked to trigger c-EB in the backward pass and allow prediction of the digit and goal in the succeeding forward pass. 

\begin{figure}[ht]
\vskip 0.2in
\begin{center}
\centerline{\includegraphics[width=0.8\columnwidth]{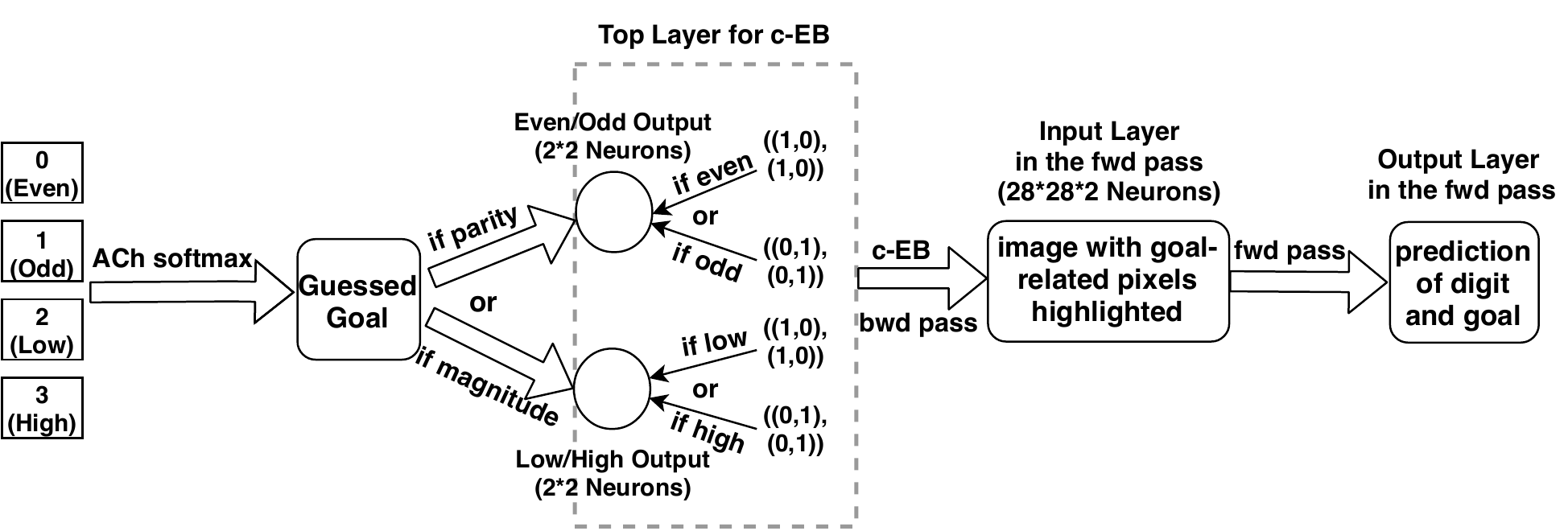}}
\caption{The neuromodulated procedure of making prediction from a guessed goal. }
\label{fig:procedure}
\end{center}
\vskip -0.2in
\end{figure}

The neuromodulated procedure of making prediction from a guessed goal is shown in Figure \ref{fig:procedure}. ACh neurons, corresponding to the four goals, were updated according to Algorithm \ref{alg:neuromod} and subjected to a softmax function to select a goal. According to the selected goal class, either two out of the four parity neurons or two out of the four magnitude neurons would be activated in the top layer to direct c-EB in the backward pass. Each of the two activated neurons corresponded to each of the two test digits, which allowed pixel highlighting of the goal-related digit on either the left or the right side of the noisy test pair image in the input layer. Finally, the goal-highlighted image went through the forward pass to predict the digit and goal for comparison with the guessed goal and the true goal digit. 

\subsubsection{Training and Test Process}
The training process consisted of incremental learning on 256 noisy MNIST pairs modified from the original MNIST training dataset \cite{lecun1998gradient}.  Every 200 training steps, 2000 test pairs of noisy MNIST digits were used to evaluate current training progress. The two digits in each training pair could have the same or opposite parity and the same or opposite high/low value(s), whereas those in each test pair all had the opposite parity and the opposite high/low values. 

During training after the forward pass, a log-softmax function, followed by a negative log likelihood function, was applied to the neurons in the output layers that each represented a digit, even parity, odd parity, high-value, or low-value. Then the sum of loss was used to calculate the gradient for each parameter in the model. At the end of each training step, a parameter update was performed based on the current gradient calculated using the Adam optimizer \cite{kingma2014adam} with a learning rate of 0.001.  

During testing, c-EB drove goal-driven perception by increasing the activity of input neurons corresponding to the goal digit and masking out the neurons corresponding to the distractor digit. In the top layer of our network (Figures 
\ref{fig:architecture} and \ref{fig:procedure}), either two out of the four parity neurons or two out of the four magnitude neurons were activated, depending on the selected goal. For example, if an ``odd'' goal was selected, the odd neuron for the left digit and the odd neuron for the right digit were both excited, whereas all other goal neurons for both digits were inhibited. In Figure \ref{fig:architecture}, a parity goal would direct c-EB by following the blue and black backward arrows to increase attention to this parity goal and its related digit, and decrease attention to the other parity goal and the two magnitude goals. This process was similar for a magnitude goal, but it would use red arrows instead of blue ones. Such c-EB generated input, with pixels highlighted for the goal digit (Figure \ref{fig:example}), then went through the forward pass again in a way similar to the training process; however, according to the goal identity, only one of the two parallel hidden layers that was related to the target goal class was used to predict the goal digit. Any step in the test process would not change the parameters of the fully trained model. 

\subsubsection{Automatic Goal Prediction}
\label{sec:automatic}
We added an online neuromodulatory model (Figure \ref{fig:procedure} and Algorithm \ref{alg:neuromod}) to the head of the network architecture in the backward pass to regulate goal selection automatically. In these experiments, the goal (with goal identities of even, odd, low-value, or high-value) had to be learned from experience. It might be noisy and rewarded with some probability (i.e., goal validity). 

Automatic goal selection was tested in 10 runs to measure the average performance. In each run, one of the four attentional goal tasks was randomly selected as the major goal, which stayed the same every 400$\pm$30 trials for 10 switches. The minor goal identity came from the same goal class as the major goal identity. For example, if the major goal was ``high'', then the minor goal became ``low'' in the same magnitude goal class; or if the major goal was ``even'', then the minor goal became ``odd'' in the same parity magnitude class. The true goal identity was set to either the major goal or the minor goal randomly according to the validity distribution per trial. The true goal digit was obtained from the labels of the test pair of noisy MNIST digits using the true goal identity. 

The goal validity values (i.e., 0.99, 0.85, and 0.70) were chosen to correspond with \cite{angela2005uncertainty}. The major goal validity was randomly chosen among the three values each time the major goal identity got switched in a run. The minor goal validity was ($1-$ major goal validity).

\section{Results}

We found the results of training and test on pairs of noisy MNIST digits to be very robust in digit and goal prediction (Section \ref{sec:results_cEB}). Section \ref{sec:results_uncertainties} shows the goal regulation performance with two valid goals within the same goal class, with one having a higher validity (major goal) and the other having a lower validity (minor goal). The major goal identity changed in the same run, whereas the major goal validity could either stay constant or change. The identity and validity of the minor goal would either change or stay constant correspondingly.

\subsection{Digit Prediction with c-EB and Noisy MNIST Pairs}
\label{sec:results_cEB}

\begin{table}[b]
\caption{Prediction for 10,000 test pairs of noisy MNIST digits.}
\label{table:digit_stats}
\vskip 0.15in
\begin{center}
\begin{small}
\begin{sc}
\begin{tabular}{lcc}
\toprule
Goal & \% Correct & \% Correct \\
Task & Digit & Goal \\
 & Prediction & Prediction \\
\midrule
Even    & 92.03 & 99.50 \\
Odd     & 91.15 & 99.75 \\
Low     & 95.39 & 99.54 \\
High    & 87.46 & 98.22 \\
\bottomrule
\end{tabular}
\end{sc}
\end{small}
\end{center}
\vskip -0.1in
\end{table}

\begin{figure}[ht]
\vskip 0.2in
\begin{center}
\centerline{\includegraphics[width=0.7\columnwidth]{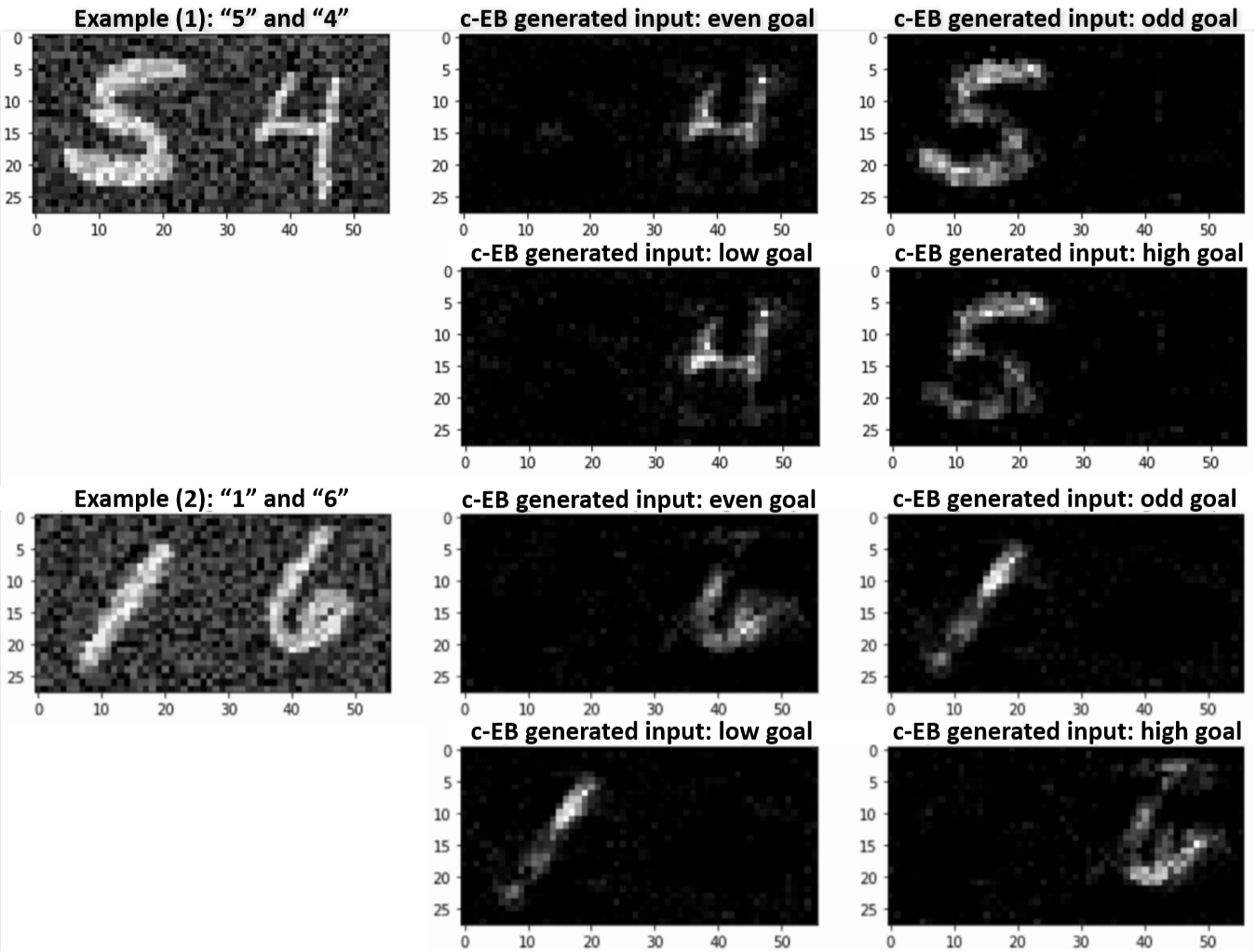}}
\caption{Two example pairs of noisy MNIST digits for testing and their c-EB highlighted results. The two digits in each test pair had the opposite goals both in the parity (even/odd) goal class and in the magnitude (high/low) goal class, whereas these limitations were not applied to the training pairs. }
\label{fig:example}
\end{center}
\vskip -0.2in
\end{figure}

The training process was carried out for 4,400 steps. The prediction performance of the fully trained model was tested on 10,000 pairs of noisy MNIST digits modified from the original MNIST test set \cite{lecun1998gradient}. Table \ref{table:digit_stats} shows the digit and goal prediction results with c-EB driven by one of the four goal tasks (i.e., even, odd, low-value, or high-value).

The goal was predicted along with the digit in the output layers for each forward pass. As shown in Table \ref{table:digit_stats}, the model predicted the goal digit correctly over 90\% of the time and predicted the goal correctly over 99\% of the time. This indicates that the goal tasks were successfully understood by the c-EB process to highlight related pixels. Although the statistics of the high-value goal task was slightly weaker than that of the other three goal tasks, the performance was still robust overall.  
Figure \ref{fig:example} shows visualization of two noisy test pairs and of their pixel highlights according to each goal task. c-EB driven by a goal went through the backward pass and excited the pixel neurons only related to the goal digit. On the irrelevant digit side, most pixel neurons were inhibited instead. Therefore, the goal digits and goal identity neurons could all be predicted correctly with high certainty in the end of the forward pass in these examples. In Example (1), ``4'' was selected when the goal was even or low , and ``5'' was selected when the goal was odd or high. Similarly in Example (2), `1'' was selected when the goal was odd or low, and `6'' was selected when the goal was even or high. However, even if two goal identities targeted the same goal digit, their highlighted pixels in c-EB visualization were not all the same. It is reasonable because our model had different output heads for the different goal tasks.

\subsection{Goal-Driven Perception with Uncertainties}
\label{sec:results_uncertainties}
 
\begin{figure}[htbp]
\vskip 0.2in
\begin{center}
\centerline{\includegraphics[width=4.7in]{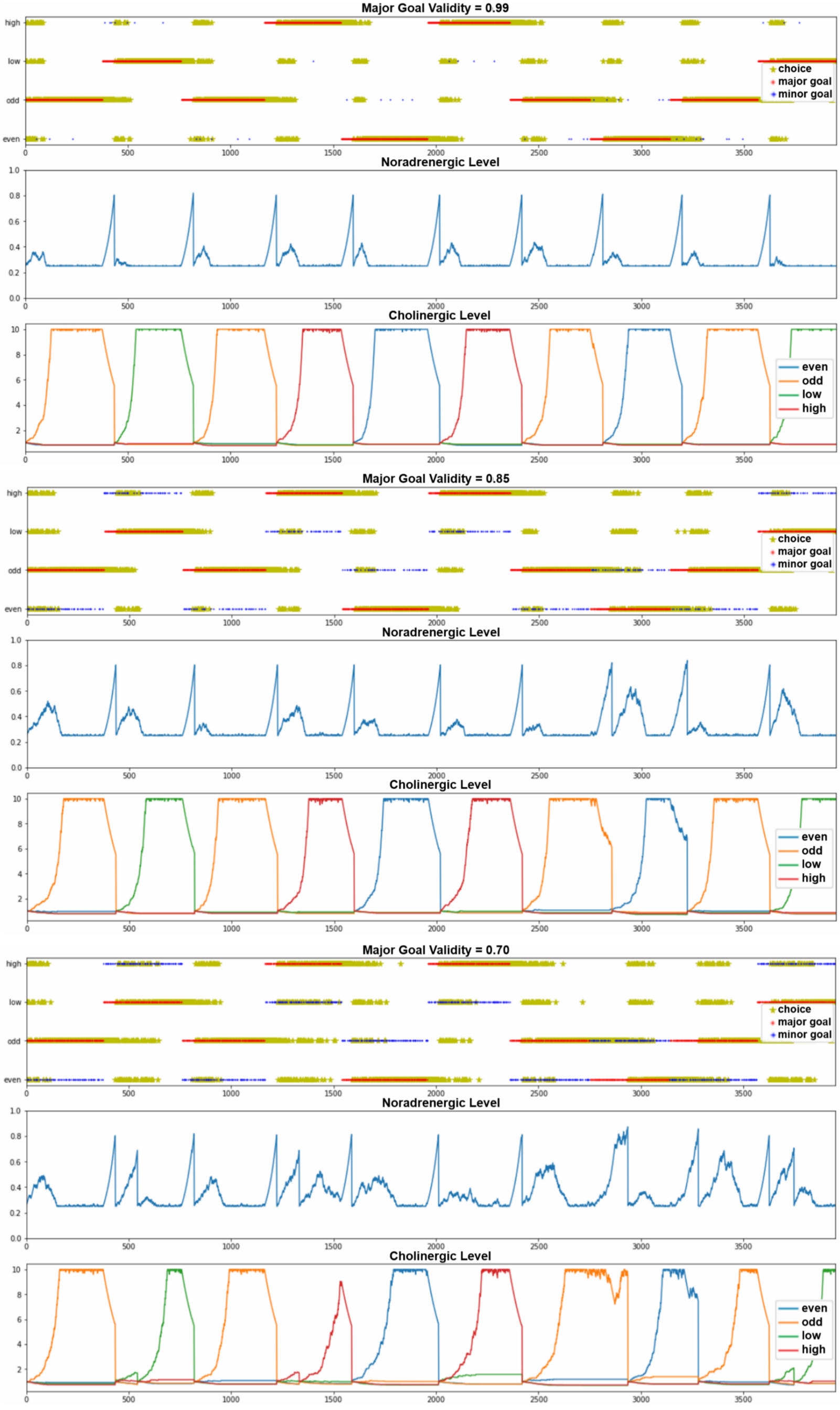}}
\caption{Visualization of neuromodulation performance for three typical runs of single major goal validities (top: 0.99, middle: 0.85, or bottom: 0.70). The major goal identity was randomly picked every 400$\pm$30 trials for 10 switches in a run. The minor goal was the other goal in the same class of the major goal. For each major goal validity, the first subplot shows guessed goal identities (in yellow dots) and true goal identities (either major goals in red dots or minor goals in blue dots). The second and third subplots show NE and ACh levels. }
\label{fig:multi_valid}
\end{center}
\vskip -0.2in
\end{figure}

Figure \ref{fig:multi_valid} shows typical runs of our neuromodulated system for three major validity settings. For each single major validity of 0.99, o.85, or 0.70, the first subplot includes the true goals (labeled as ``major goals'' and ``minor goals'') and ACh-guessed goals (labeled as ``choice''); the second and third subplots show NE and ACh levels. Note that the ACh neuron corresponding to a major goal quickly increased driving attention to the most likely goal, as well as suppressing attention to distractors. In cases where the major goal validity was low, the ACh neuron corresponding to the minor goal was also activated, resulting in more exploration. However, the prediction during exploration tended to remain in the same goal class. When there was a change in the goal identity, the NE neuron quickly recognized the change and responded with spike of activity. This caused the network to reset, and a short period of exploration before the system found the new goal identity. Lower major goal validity led to longer exploration, especially after a goal identity switch, as well as more frequent NE bursts. 

\begin{figure}[ht!]
\vskip 0.2in
\begin{center}
\centerline{\includegraphics[width=5.0in]{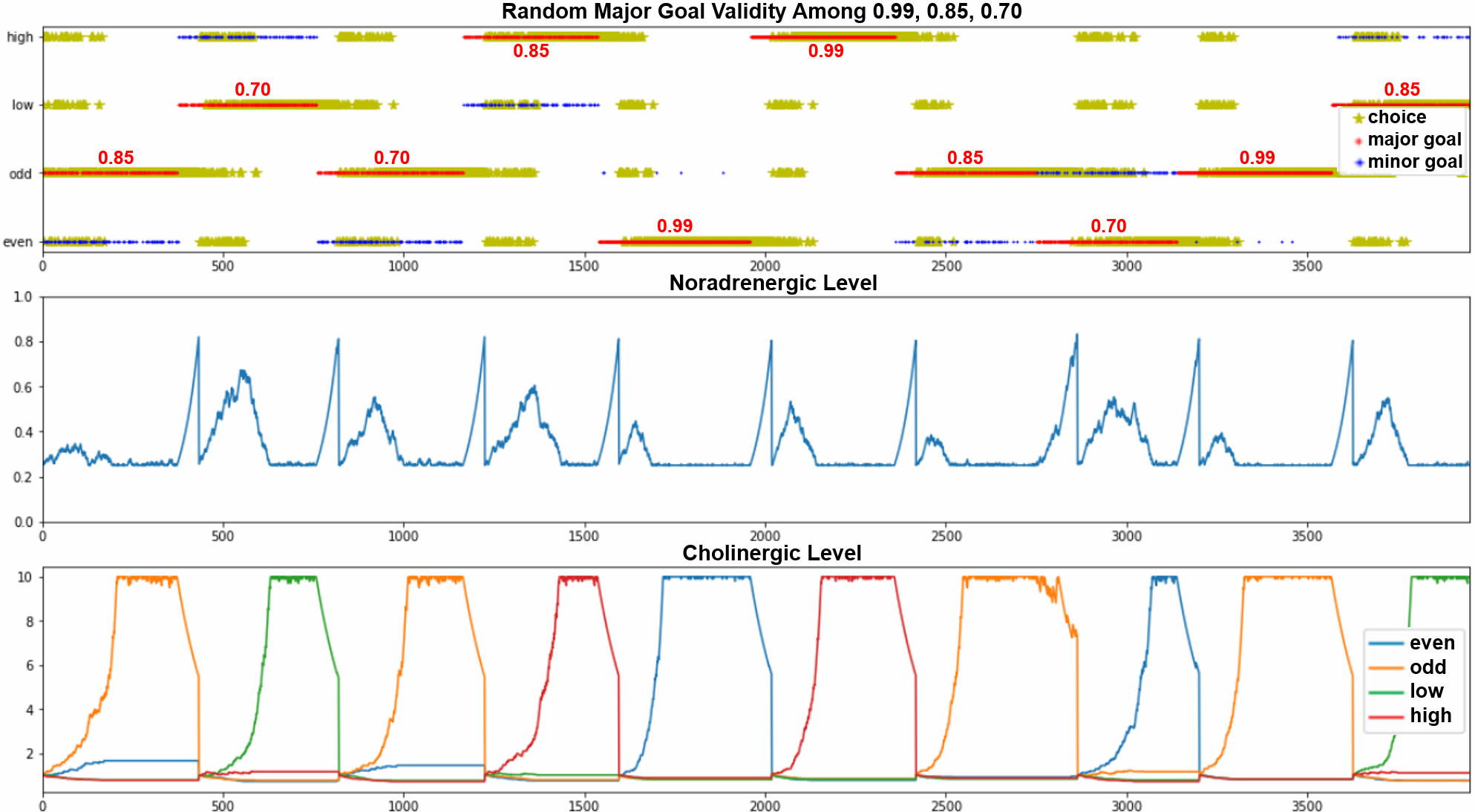}}
\caption{Visualization of neuromodulation performance for a typical run of where major goal validity switched among 0.99, 0.85, and 0.70.}
\label{fig:multi_random}
\end{center}
\vskip -0.2in
\end{figure}

We also ran experiments where the goal validity could change during the run. Figure \ref{fig:multi_random} shows the performance of a typical run with random switching among three major goal validity options, 0.99, 0.85, and 0.70. Similar to Figure \ref{fig:multi_valid}, the system in this setting still focused more on the major goal. When the minor goals occurred more frequently, the NE neuron fired phasically more frequently; meanwhile, the major goal's ACh neuron was less, but the minor goal's ACh neuron had a low level of excitation. Because both the goal validity and goal identity changed during a run, the exploration period lasted longer with a major goal validity of 0.85 or 0.70. However, this also led to higher prediction accuracy of the minor goal.

\begin{table}[ht]
\caption{Average neuromodulation performance over 10 runs for each of the four goal validity settings. The first three rows of data correspond with the first experiment of one major goal validity. The final row relates to the second experiment of randomly switched goal validity. $p\textrm{\_}valid$ means the major goal validity, and $(1-p\textrm{\_}valid)$ means the minor goal validity. In each run, the major goal was randomly picked every 400$\pm$30 trials for 10 switches. The minor goal was selected from the same goal class. } 
\label{table:multi_valid}
\vskip 0.15in
\begin{center}
\begin{small}
\begin{sc}
\begin{tabular}{c|c|c|c|c|c}
\toprule
 Major & Minor & \% Correct & \% Correct & \% Incorrect & \% Incorrect \\
Goal & Goal & Major & Minor & ACh Softmax & c-EB  \\
Validity & Validity & Goal & Goal & Prediction & Prediction \\
\midrule
0.99 & 0.01 & 67.0 & 0.1 & 26.0 & 6.9\\
0.85 & 0.15 & 54.0 & 1.3 & 38.9 & 5.8\\
0.70 & 0.30 & 37.4 & 4.8 & 53.3 & 4.5\\
p\textrm{\_}valid & 1-p\textrm{\_}valid & 49.5 & 12.4 & 31.7 & 6.4\\
\bottomrule
\end{tabular}
\end{sc}
\end{small}
\end{center}
\vskip -0.1in
\end{table}

Table \ref{table:multi_valid} shows the average goal prediction performance over 10 runs for each validity setting. The third and fourth columns refer to the percent of trials at which the goal digit prediction was correct. The fifth column refers to the percent of incorrect goal guess from ACh-related softmax selection. The sixth column refers to the percent of incorrect digit prediction with c-EB driven by the guessed goal, when the ACh-guessed goal already matched the true goal. The first three rows provide average statistics for runs at which a single goal validity was tested (see also Figure \ref{fig:multi_valid}), and the last row corresponds with runs at which the goal validity could change randomly among three options during a run (see also Figure \ref{fig:multi_random}).

\section{Discussion}

In this paper, we showed how contrastive excitatory backpropagation (c-EB) could be used in a goal-driven perception task. Goals are often unknown and need to be discovered. Therefore, we extended a biologically plausible model of neuromodulation to learn goals and to rapidly adapt when they change. The c-EB algorithm was modified to support multiple goals. This allowed attention to be diverted to the chosen goal, while ignoring distractors.

 Yu and Dayan \cite{angela2005uncertainty} proposed a Bayesian model of neuromodulation in which the cholinergic system tracked expected uncertainty and the noradrenergic system tracked unexpected uncertainty. The present paper advances this work in two ways to support goal-driven perception: 1) The Bayesian model was recast as a neural model to make it compatible with neural networks. The neuromodulators were implemented as a neural network layer to drive attention toward a goal digit and divert attention away from distractors. 2) A neural network reset was implemented to rapidly re-adapt when a goal changes. Empirical evidence suggests that the noradrenergic system facilitates this reset in the brain \cite{bouret2005network,Grella445}. In our experiments, the noradrenergic system rapidly recognized a change in the goal contingency, and drove a reset of cholinergic activity. This caused the neural network to quickly explore new goals.
 
 In the real world, goals are often uncertain. In our experiments, the goal validity (i.e., probability of a goal being rewarded) ranged from 0.99 to 0.85 to 0.70, and the system had to respond by either choosing the most likely goal or exploring alternative goals. Furthermore, the experimental design had a hierarchy of goals. For example, the goal would be to attend to the parity goal class and the sub-goal might be to reward odd digits 70\% of the time and even digits 30\% of the time. Interestingly, the neural network would often stay within a goal class (i.e., choose parity and not magnitude).
 
 Exploring options, rather than always choosing the most likely goal, is known as probability matching behavior \cite{wozny2010}. Similar to the results presented here, humans tend to underselect the most rewarding \cite{craig2016}. Such behavior may be due to feature exploration, as subjects test hypotheses by switching between the features before deciding upon their most rewarding goal. In rodent studies, it has been shown that rats will seek uncertainty, and that this uncertainty seeking is governed by the cholinergic system \cite{naude2016}. These uncertainty seeking strategies that appear in natural systems may be advantageous for artificial systems that are deployed in dynamic environments.


Top-down task-driven attention is an important mechanism for efficient visual search in humans and artificial systems \cite{baluch2011}. Many computational models of attention have been proposed and implemented to either explain top-down attention or develop an application inspired by these mechanisms \cite{tsotsos2015}. Of particular interest are attentional systems that can leverage the power of CNNs. In these cases attentional information can propagate backwards, highlighting the features of a given goal \cite{zhang2018top,zhou2016learning}. Similar to the effect of the cholinergic system to increment attention to a goal and decrement attention to distractors \cite{baxter1999cognitive}, Zhang and colleagues \cite{zhang2018top} proposed an excitatory backpropagation mechanism with a contrastive top-down signal to enhance the perception of goal features. Similarly, Zhou and colleagues \cite{zhou2016learning} proposed a technique called Class Activation Mapping (CAM) for identifying regions in an attentional map. Selvaraju and colleagues \cite{selvaraju2017grad} proposed Gradient-weighted Class Activation Mapping (Grad-CAM) to highlight regions of interest and generate visual explanations. Their model could be applied to any CNN with no re-training. Similarly, our proposed model can work with any CNN. Moreover, our model replaces the Winner-Take-All mechanism or rigid probabilistic methods, with a flexible and adaptable layer based on neurobiological neuromodulation.

In the present work, the goal classes were known, and the system predicted the appropriate goal given a goal identity and goal validity. However, the system might need to adapt to new goals or new goal classes. Adding multiple heads to the output layer of the network is one way this could be handled. This would not require retraining the stimuli (e.g., digits), but some additional training for the new goal classes. However, the architecture might be more scalable with a single head that is trained to an arbitrary number of goals.  This will be explored in future iterations of our work.

The choice of c-EB for a top-down attentional mechanism was motivated by its similarity to cholinergic system and its affect on top-down attention. However, as mentioned above, we believe that the proposed system could also work with other state-of-the-art attentional mechanisms, including the CAM \cite{zhou2016learning} and its more general variation Grad-CAM \cite{selvaraju2017grad}. As long as the neural network structure can add an additional neuromodulation layer, and there is some means to flow goal information from the top to lower layers, our neuromodulatory goal-driven perception system should be compatible.

\section{Conclusions}
In this paper, we introduce a model of cholinergic and noradrenergic neuromodulation to perform goal-driven perception. The proposed network architecture discovers goals using online learning, and highlights the stimulus features corresponding to the goal. Moreover, the proposed system rapidly adapts when goal contingencies change. This neurobiologically inspired model can be applied to other problem domains and other top-down attentional networks.

\section*{Acknowledgements}

This material is based upon work supported by the United States Air Force and DARPA under Contract No. FA8750-18-C-0103. Any opinions, findings and conclusions or recommendations expressed in this material are those of the author(s) and do not necessarily reflect the views of the United States Air Force and DARPA.

\end{document}